# BuildMapper: A Fully Learnable Framework for Vectorized Building Contour Extraction

Shiqing Wei, Tao Zhang, Shunping Ji*, Muying Luo, Jianya Gong

a *School of Remote Sensing and Information Engineering, Wuhan University, 129 Luoyu Road, Wuhan 430079, China*

* Corresponding author

E-mall addresses: wei_sq@whu.edu.cn (S. Wei), zhang_tao@whu.edu.cn (T. Zhang), jishunping@whu.edu.cn (S. Ji), luomuying@whu.edu.cn (M. Luo), Gongjy@whu.edu.cn (J. Gong)

**Abstract**

Deep learning based methods have significantly boosted the study of automatic building extraction from remote sensing images. However, delineating vectorized and regular building contours like a human does remains very challenging, due to the difficulty of the methodology, the diversity of building structures, and the imperfect imaging conditions. In this paper, we propose the first end-to-end learnable building contour extraction framework, named BuildMapper, which can directly and efficiently delineate building polygons just as a human does. BuildMapper consists of two main components: 1) a contour initialization module that generates initial building contours; and 2) a contour evolution module that performs both contour vertex deformation and reduction, which removes the need for complex empirical post-processing used in existing methods. In both components, we provide new ideas, including a learnable contour initialization method to replace the empirical methods, dynamic predicted and ground truth vertex pairing for the static vertex correspondence problem, and a lightweight encoder for vertex information extraction and aggregation, which benefit a general contour-based method; and a well-designed vertex classification head for building corner vertices detection, which casts light on direct structured building contour extraction. We also built a suitable large-scale building dataset, the WHU-Mix (vector) building dataset, to benefit the study of contour-based building extraction methods. The extensive experiments conducted on the WHU-Mix (vector) dataset, the WHU dataset, and the CrowdAI dataset verified that BuildMapper can achieve a state-of-the-art performance, with a higher mask average precision (AP) and boundary AP than both segmentation-based and contour-based methods. We also confirmed that more than 60.0/50.8% of the building polygons predicted by BuildMapper in the WHU-Mix (vector) test sets I/II, 84.2% in the WHU building test set, and 68.3% in the CrowdAI test set are on par with the manual delineation level.

**Keywords:** Building contour delineation, Instance segmentation, Contour-based method, Deep learning, Remote sensing images

## 1 Introduction

For many years, researchers have been engaged in developing an automated method that can replace a human to directly draw the structured vector format contours of individual buildings (Fig. 1(b)), which play an essential role in GIS production, urban planning, population density estimation, energy supply, and disaster management [1-3]. Undoubtedly, this is extremely challenging, not only due to the difficulty of how to design such a sophisticated and smart algorithm, but also the challenges that come from imperfect imaging conditions, diverse building architectures, and background complexity. In this paper, we propose an end-to-end and fully learnable method that can directly delineate the regular building contours in vector format. In this section, we briefly review the development history of building extraction and the ideas that have inspired us, and then introduce the proposed approach and summarize the contributions.

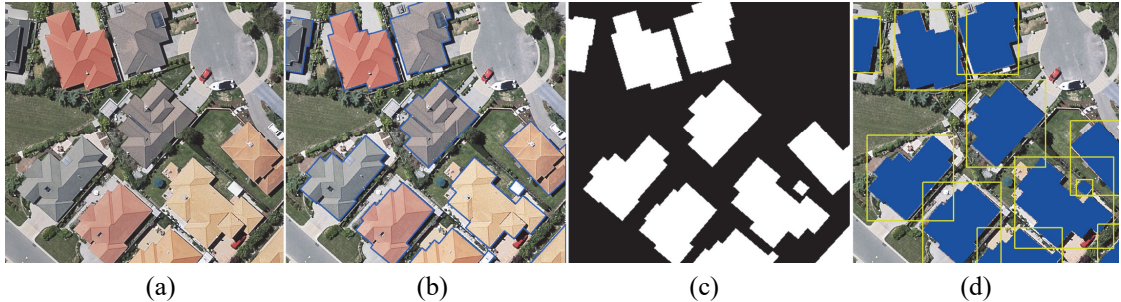

Fig. 1. Examples of building extraction results by different methods in ideal cases. (a) Input image. (b) manual delineation, which we pursue with an end-to-end contour-based method. (c) Semantic segmentation. (d) Instance segmentation.

Over the past decades, the conventional methods have been used to extract buildings by the use of textures, lines, shadows, and other empirically designed features [4-6]. However, this kind of empirical method cannot generalize to varied imaging conditions, different building shapes, and complex backgrounds. Recent deep learning based techniques have significantly improved the accuracy level of building extraction from remote sensing images. From the early convolutional neural networks (CNNs) [7] to fully convolutional networks (FCNs) [8], most deep learning based methods formulate building extraction as a semantic segmentation task [9-16], which involves partitioning an image into regions with different categories (e.g., building or background, as shown in Fig. 1(c)). Clearly, semantic segmentation alone cannot discriminate individual buildings.

Instance segmentation methods such as Mask R-CNN [17], SOLO [18], YOLACT [19], and PANet [20] can be applied to predict both individual building locations by bounding boxes and a building mask (Fig.1(d)). However, a bounding box predicted by these methods can contain additional instances that interfere with the prediction of the current building mask [21]. For example, the biggest box in the lower-right corner of Fig. 1(d) contains parts of other buildings.

Both semantic segmentation and instance segmentation methods produce pixel-level segmentation maps, but the building boundaries they produce are usually zigzagged, irregular, broken, and far from the manual delineation level. Moreover, the results require extensive post-processing. For example, a knowledge-based regularization algorithm [22-25] can be used to transform segmentation maps into vector format building contours.

A few other studies [26-29] have considered the instance segmentation problem as contour regression, i.e., regressing the vertex coordinates of a contour (in other words, a polygon represented by a series of discrete vertices). The contour-based methods are theoretically advantageous in efficiency since they straightforwardly regress the polygon coordinates, compared to semantic/instance segmentation with a pixelwise operation, and have the potential to get rid of the need for post-processing operations such as raster-to-vector conversion and regularization.

Our idea in this study was inspired by the contour-based methods. Therefore, in the following, we briefly review the development of this type of method, and explain the difficulties encountered when applying this type of method to structured building contour extraction. The traditional contour-based methods [30, 31] use an active contour model to extract the contours of the objects of interest by optimizing an empirically designed energy function. Recent methods [32, 33] have combined an active contour model with a CNN, which improves the robustness of the model. In more recent studies, researchers have realized contour extraction with a unified deep neural network framework. For example, a number of researchers [34-38] have used a recurrent neural network (RNN) [39] to predict the corner vertices of buildings in the clockwise direction. However, this type of method often suffers from corner vertex loss and incorrect vertices. CNN-based methods are currently the mainstream contour-based methods. The general one-stage contour-based CNN methods, such as PolarMask [40], PolarMask++ [41], and LSNet [42], are highly efficient, but usually obtain only rough contours of objects as they use only deep features of the instance centers. Most of the contour-based methods utilize a two-stage pipeline: initial contour generation and contour evolution. The shapes of the initial contours used in the current contour-based methods are generally hand-designed, such as ellipses for Curve GCN [26], rectangles for DANCE [28], and octagons for Deep Snake [27]. The difference between the initial contours and the ground-truth polygon can dramatically increase the learning difficulty of a network, and challenge the subsequent contour evolution. The PolyTransform [43] and PolygonCNN [44] methods use the segmentation result as the initial contours, but they are dependent on a well-predicted segmentation map, and the whole process is complex and inefficient. The other obvious drawback of the modern contour-based methods is the supervision strategy, in that they supervise the network by computing the location difference between the pre-paired predicted vertices and ground-truth contour vertices. However, the corresponding relationship between the ground-truth vertices and predicted vertices is much more complex than in pixel-level segmentation. In most of the current contour-based methods [26-28], they apply a preset static correspondence, regardless of the dynamic adjustment of the predicted contour vertices. For example, a predicted vertex that is already on the ground-truth boundary but far from its correspondence vertex has to be continuously adjusted, uselessly and even harmfully.

In addition to the common problems in contour initialization and vertex-based supervision, we found that the modern contour-based methods, such as Curve GCN and Deep Snake, tend to predict over-smooth contours that are far from regular building shapes. The reasons for this are twofold. Firstly, the existing contour-based methods associate the features of multiple adjacent vertices, e.g., nine vertices in [26, 27, 45], for offset prediction of the current vertex. This approach enriches the local information, but it also weakens the current vertex information, leading to over-smooth results and increased computational cost. Secondly, these methods predict the contours with a given fixed number $N$ of vertices for each building, where $N$ (e.g., 64) is much larger than the number of building corner vertices, which introduces over-smoothing. Furthermore, these methods cannot directly locate and draw the corner vertices of a building as a human does. The CLP-CNN method [29] attempts to remove predicted vertices that are far from the building corner vertices. However, this approach is fairly rough and can lead to incomplete polygons in some cases.

Currently, an end-to-end building contour extraction method that directly outputs structured polygons consisting of building corner vertices is lacking. The recent contour-based methods, like the semantic and instance segmentation methods, have not got rid of the need for independent post-processing including polygon simplification and regularization. However, a simple polygon simplification algorithm (e.g., the Douglas-Peucker algorithm [46]) cannot directly output satisfactory structured building contours. The empirical regularization algorithms [22-25] also have undeniable drawbacks: 1) they are not only heavily dependent on the quality of the predicted building segmentation/contour results, but they also rely on many parameter settings, which leads to a poor generalization ability; 2) most of the methods cannot be applied to buildings with arbitrary shapes; and 3) they often lead to low automation and efficiency.

It is also worth noting that the lack of special building datasets limits the development of contour-based building extraction methods. Currently, the annotations of the open-source building datasets are presented in different formats, mainly raster format such as TIFF and PNG, but vector format is really what is required by the contour-based methods. In addition, label errors, including misalignments, omissions, and mislabeling, which are common in OpenStreetMap [47], the Ordnance Survey datasets [48], the SpaceNet dataset [49], the Crowd AI dataset [50], and the Inria dataset [51], reported in [52], decrease the training effect and prediction quality of a model.

In this paper, we tackle all of the above-mentioned problems, including the common initialization and supervision problems in the contour-based methods, the over-smoothing problem in building extraction, the problems of using empirical post-processing, and also the problems related to the datasets. We propose new ideas for achieving the direct extraction of structured and vector format building contours. It should be noted that this paper is an extension of our previous conference work [53], in which we solved the initialization and supervision problems by presenting a learnable contour initialization method to replace the hand-designed initial contours in the previous studies, and proposed a dynamic matching loss for network supervision to solve the static vertex correspondence problem. Other brand-new contributions include:

1) We propose an end-to-end contour-based learnable framework for high-quality building contour extraction named BuildMapper, which is simple in structure, high in accuracy, and extremely fast. The proposed framework can directly output regular building contours consisting of corner vertices. To the best of our knowledge, this is the first end-to-end method that does not require any post-processing.

2) We propose a novel learnable contour evolution method that performs both contour vertex deformation and reduction, and removes the need for empirical post-processing. Vertex information extraction and aggregation is realized with a lightweight feature encoding structure, and a well-designed vertex classification head is introduced to select corner vertices. This head assigns a best-match vertex to each ground-truth corner vertex, and removes the other vertices.

3) To tackle the problems of the lack of datasets for contour-based vector format building extraction, we built a new large-scale high-quality building dataset based on the WHU-Mix building dataset [52], which we call the WHU-Mix (vector) dataset, to advance the tremendous potential of the contour-based building extraction methods.

4) The experimental results obtained on the WHU-Mix (vector), WHU, and CrowdAI datasets demonstrated the state-of-the-art performance of BuildMapper and its significant speed advantage when compared to both instance segmentation and contour-based methods. A new automation level on par with the manual delineation level is also reported.

The remainder of this paper is structured as follows. Section 2 introduces the details of BuildMapper. In Section 3, the details of the experimental setup and the datasets used are provided. We present the results of the experiments and discuss optional designs in Section 4. Finally, we draw our conclusions in Section 5.

## 2 Methodology

The proposed framework integrates individual building localization and structured building contour extraction into an end-to-end simple and high-efficiency deep learning network, which we have named BuildMapper. As illustrated in Fig. 2, the core idea of BuildMapper includes a learnable contour initialization and a contour evolution module. Given an image, the initialization module adaptively generates a suitable initial polygon for each building instance, based on the deep feature map of the input image. The contour evolution module then evolves the polygon vertices twice from the initial position to approach the real building boundary, and finally eliminates non-corner vertices to form manual-delineation-level polygons represented by corner vertices. Details of BuildMapper are provided below.

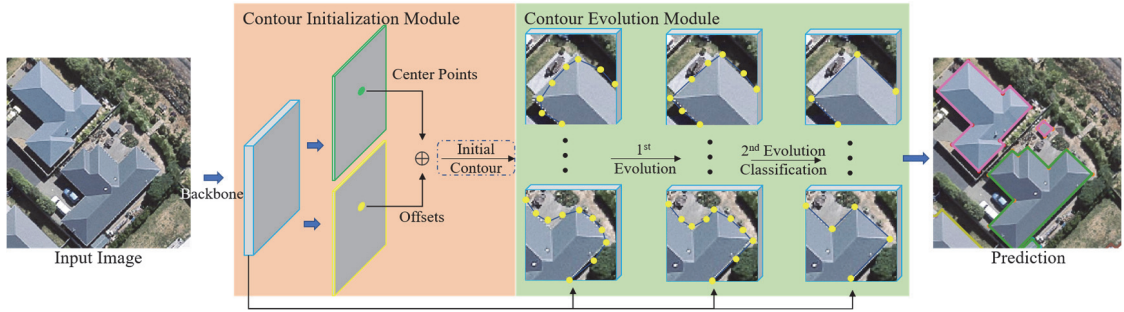

Fig. 2. The structure of the proposed BuildMapper. A given image is processed by the contour initialization module, which generates an initial polygon for each building instance, and then the contour evolution module, which adjusts the polygon vertices and eliminates redundancy vertices by a classification head, to output regular building polygons.

## 2.1  Preliminary work

A building contour in an image can be delineated as a polygon. The polygon can be presented by the corner vertices of the building in a clockwise (or counterclockwise) sequence. The $i$-th corner vertex is denoted as $p_i^c$, and the ground-truth building contour is denoted as $gt^c = \{p_i^c\}_{i=1}^M$, where $M$ is the number of corner vertices. To facilitate a contour-based supervised learning framework, we need to interpolate the polygon with more vertices. In the proposed approach, we set multiple anchors in the ground-truth contours of the buildings to reduce the learning difficulty of the proposed network. Firstly, we insert $s$ control vertices into the building contour with $s$ fixed directions (relative to the center point of the building contour), as shown by the yellow points in Fig. 3(b). The building contour segment between every two adjacent control vertices is then densely resampled to the same number of vertices (see Fig. 3(c)). In the experiments, $s$ was set to 4, i.e., the control vertices were inserted in the *top*, *bottom*, *left*, and *right* directions of the building, respectively. After this, we obtain the ground-truth building contour $gt = \{p_i\}_{i=1}^N$ with $N$ vertices. In this study, $N$ was set to 64. We set the control vertex in the *top* direction as the starting vertex of $gt$, as shown in Fig. 3(c). Meanwhile, we define the predicted building contour as $pre = \{\hat{p}_i\}_{i=1}^N$. In this way, we can design appropriate vertex pairing strategies for $gt$ and $pre$ to supervise the building contour extraction network. Finally, at the back-end of the network, the $N$ boundary vertices are automatically reduced to $M$ corner vertices, to obtain a polygon that approaches the ground truth.

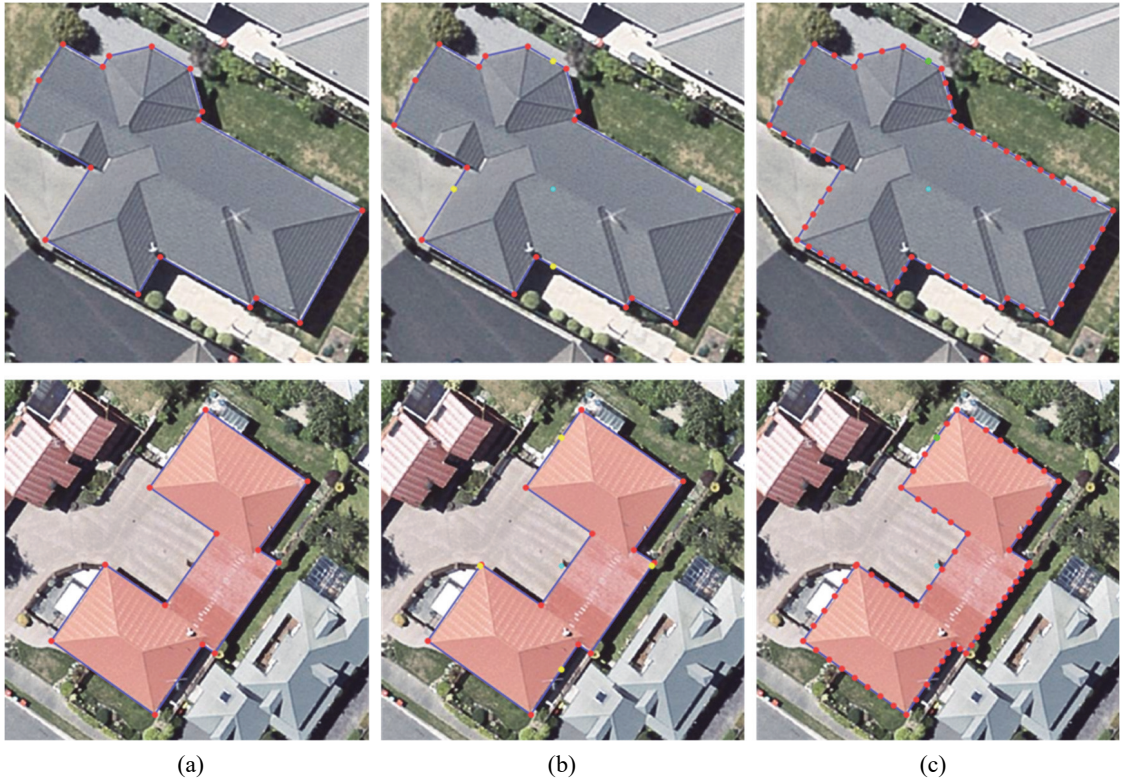

(a)                (b)                (c)

Fig. 3. Example of dense resampling of building contours, where the cyan dot is the center point of the bounding box, the inserted control points are yellow, the start point of the polygon is green, and the other points are red. (a) Original polygon. (b) Control vertex insertion. (c) Densified polygon.

## 2.2 Contour initialization module

The pipeline of the contour initialization module is shown in Fig. 4. The contour initialization module first predicts the position (center point) of each building instance in the image, and then regresses the coordinates of the $N$ contour vertices based on the deep features of the current building instance. The contour initialization module contains a backbone convolutional network, a building center point prediction head, and an offset prediction head.

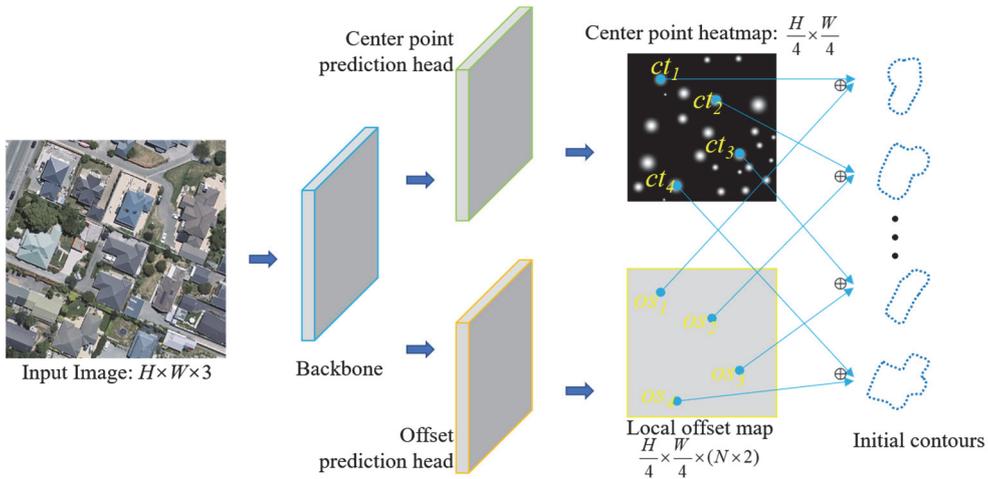

Fig. 4. Pipeline for the contour initialization module in BuildMapper. $ct$ represents the coordinates of the center point of each building on the image. $os$ represents the corresponding offsets. $\oplus$ is addition operation for obtaining the initial contour of each building.

### 2.2.1 Backbone network

We use a modified deep layer aggregation (DLA) network [54] as the CNN backbone to extract high-level features of the input images, as in CenterNet [55]. The DLA network iteratively merges the feature information by hierarchical skip connections, allowing for better accuracy and fewer parameters. The skip connection from low-resolution feature maps to original-scale output is enhanced by deformable convolution [56]. Give a color input image $I \in \mathbb{R}^{W \times H \times 3}$ of width $W$ and height $H$, the output of the CNN backbone network is $F \in \mathbb{R}^{\frac{W}{4} \times \frac{H}{4} \times C}$, where $C$ is the number of feature map channels. In this study, we set $C = 64$. In the proposed framework, the feature map $F$ is shared with the center point prediction head, the offset prediction head, and the subsequent contour evolution module.

### 2.2.2 Buildings as points

Empirically, spatial location can be used as a cue to distinguish object instances, as objects are placed in different locations [57]. Thus, individual building instances in the image can be roughly represented by their center point. In this way, the building object detection task can be transformed into a standard key point estimation problem. The aim of the center point prediction head is to produce a key point heatmap $\hat{Y} \in [0,1]^{\frac{w}{4} \times \frac{h}{4}}$. The center point prediction head includes a 3×3 convolution, a rectified linear unit (ReLU), another 1×1 convolution, and a sigmoid activation function. The peaks in heatmap $\hat{Y}$ correspond to building centers.

Clearly, the center points of the buildings in the image are very few compared to the other pixels. In order to reduce the imbalance effect of the positive and negative samples, the center point prediction head is trained using focal loss [58]. The parameter settings of the focal loss are the same as those of CenterNet [55]. The center point loss $L_{ct}$ is defined as shown in Eq. (1):

$$L_{ct} = -\frac{1}{M} \sum \begin{cases} (1-\hat{Y})^\alpha \log(\hat{Y}), & \text{if } Y = 1 \\ (1-\hat{Y})^\beta (\hat{Y})^\alpha \log(1-\hat{Y}), & \text{otherwise} \end{cases} \quad (1)$$

where $Y$ is the ground truth of the key point heatmap, $\alpha$ and $\beta$ are hyperparameters set to 2 and 4, and $M$ is the number of key points.

During the training phase, the ground truth building center points are used for offset prediction. At the inference time, we first extract the peaks in the heatmap $\hat{Y}$. A max-pooling operation of kernel size 3×3 is used as non-maximum suppression in their neighborhoods. Pixel locations with values greater than the threshold $T$ (e.g., $T$=0.2) among the top $K$ (e.g., $K$=200) peaks are selected as the predicted building center point.

### 2.2.3 From points to initial contours

Inspired by PolarMask [40], features at the object center point that contain a wealth of information can be used to predict the contour of the current object. Given the coordinates of the building center point and the contour, we can calculate the offset of each contour vertex with respect to the center point. The offset of each vertex can be denoted as $\{(\Delta x_i, \Delta y_i) | i = 1, 2, \ldots, N\}$. The offset prediction head produces a local offset map $O \in \mathbb{R}^{\frac{W}{4} \times \frac{H}{4} \times N \times 2}$, and extracts the corresponding contour offsets in $O$. Finally, the initial contour $py^{init}$ of the building is obtained by adding the predicted $N$ offsets to the center point coordinates $(x_{ct}, y_{ct})$, as shown in Eq. (2):

$$py^{init} = (x_{ct}, y_{ct}) + \gamma \times \{(\Delta x_i, \Delta y_i) | i = 1, 2, \ldots, N\} \quad (2)$$

where $\gamma$ is the expansion factor, which we set to 10 in all the experiments described in this paper. The structure of the offset prediction head is similar to that of the center point prediction head, including a 3×3 convolution, a ReLU, and another 1×1 convolution. The predicted initial contour is then sent to the contour evolution module for further adjustment.

Note that the building contours resampled with multi-directional anchors have the same number of contour vertices between adjacent anchor vertices. Therefore, the network only needs to predict a fixed number of initial contour offsets in each directional interval, which greatly reduces the training difficulty of the module, which is a problem that is commonly encountered by the previous contour-based methods.

Since the direction of the starting vertex of *gt* is fixed (*top* direction), we can supervise the offset prediction head using smooth $L_1$ loss [59]. This loss function can ensure that the gradient value of the network is not too large when the difference between the predicted initial contour and ground truth is very large, and is small enough when the difference is small. The initial contour loss calculation can be defined as:

$$L_{init} = \frac{1}{N}\sum_{i=1}^{N} l_1(py_i^{init} - gt_i) \tag{3}$$

where $l_1$ represents the smooth $L_1$ loss function, for which we refer the reader to [58] for further details.

## 2.3 Contour evolution module

The initial contour of the building inferred only from the feature information of the center point is rough, and thus needs to be further adjusted by the contour evolution module, as shown in Fig. 5. The contour evolution module includes two functions: contour vertex coordinate adjustment and redundant vertex removal. It takes both the coordinates and feature information of the contour vertices as input, iteratively adjusts the vertices to the suitable positions on the building boundary, and removes the redundant vertices.

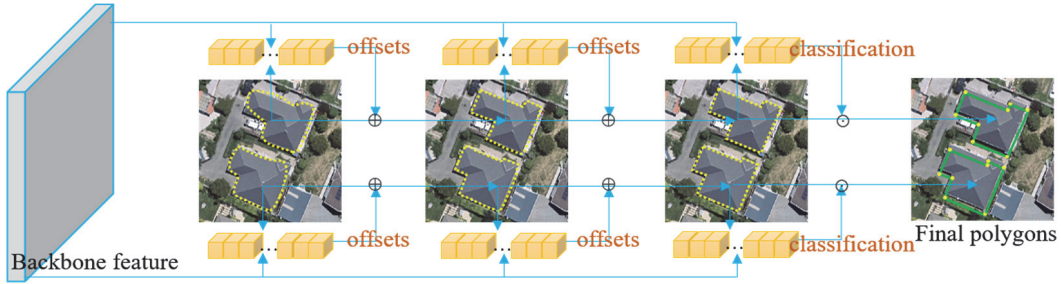

Fig. 5. Pipeline for contour evolution module in BuildMapper. ⊕ means adding the predicted offsets to the current polygon coordinates to obtain the adjusted building contour. ⊙ denotes redundant vertex elimination according to the vertex classification confidence.

### 2.3.1 Contour vertex coordinate adjustment

Given the initial contours, we first construct feature vectors for each vertex. The feature vector $f_i$ of each vertex $i$ in the contour evolution module is a concatenation of the vertex's relative coordinates and the feature vector extracted from the corresponding location in the feature map $F$. Hence, $f_i$ = concat{$F_i$, $x_i^{rel}$, $y_i^{rel}$}. $F_i$ is obtained by bilinear interpolation. $x_i^{rel}$ and $y_i^{rel}$ are the regularized coordinates of the current vertex, which are obtained by the use of Eq. (4):

$$\begin{aligned} x_i^{rel} &= (x_i - x_{ct})/(x_{max} - x_{min}) \\ y_i^{rel} &= (y_i - y_{ct})/(y_{max} - y_{min}) \end{aligned} \tag{4}$$

where ($x_{min}$, $x_{max}$), ($y_{min}$, $y_{max}$) denote the minimum and maximum values of the *x* and *y* coordinates of the current building contour vertices, respectively. The regularized coordinate values range from −0.5 to 0.5, which are more stable than the absolute coordinate values. For each building contour, the dimensionality of its feature vectors is $N×(64+2)$. Taking $f_i$ as input, the vertex adjustment head outputs the predicted offset $[\Delta x_i, \Delta y_i]$ of each vertex coordinate, which can then be updated by $[x_i', y_i'] = [x_i, y_i] + [\Delta x_i, \Delta y_i]$. In this way, the contour completes a round of evolution. In all the experiments, we set the number of iterations *iter* to 2.

### 2.3.2 Contour vertex reduction

A contour inferred by BuildMapper is a fixed-size set of $N$ predictions, where $N$ is set to be significantly larger than the typical number $M$ of corner vertices in a building contour. Contour vertex reduction classifies the predicted vertices into valid and invalid. A simple rule such as calculating the minimum distance between the predicted contour vertex and the building corner vertex can easily lead to incomplete building shapes when the predicted contour is not perfect. In fact, we can construct an optimal bipartite matching between the predicted contour $pre = \{\hat{p}_i\}_{i=1}^{N}$ and the corresponding ground-truth contour $gt^c = \{p_i^c\}_{i=1}^{M}$, to determine which vertices can be eliminated. To find the optimal bipartite matching between *pre* and $gt^c$, we search for a permutation with the lowest matching cost that assigns a matching vertex in *pre* for each vertex in $gt^c$, considering both the classification confidence score and the distance between the ground-truth vertices and predicted vertices.

The Hungarian algorithm [60, 61] can efficiently compute this optimal allocation. To derive the predicted contour vertex indexed as $\sigma(i)$ that is matched to point $i$ in $gt^c$, we calculate the probability of classification, denoted as $c_{\sigma(i)}$, which is obtained by applying a softmax function on the classification head. The distance between the ground-truth vertices and the predicted vertices is denoted as $\hat{p}_{\sigma(i)} - p_i^c$. Thus, by minimizing the pair-wise matching cost $L_{match}(p_i^c, \hat{p}_{\sigma(i)})$, as shown in Eq. (5), $\sigma(i)$ can be obtained.

$$L_{match}(p_i^c, \hat{p}_{\sigma(i)}) = -c_{\sigma(i)} + \delta \times (\hat{p}_{\sigma(i)} - p_i^c) \tag{5}$$

In the experiments, we set $\delta$ to 5. During training, the vertices in *pre* that match the vertices in $gt^c$ are set as valid (label value 1), and the vertices that do not match are set as invalid (label value 0). We can then use a negative log-likelihood loss function to supervise the contour vertex classification head, as shown in Eq. (6). Since the number of vertices of *pre* is much larger than $gt^c$, the number of invalid vertices is also much larger than the number of valid vertices. Therefore, in practice, we down-weight the log-probability term by a factor of 10 to account for the class imbalance. At the inference time, the vertices with a confidence less than $t$ in the predicted contour are classified as invalid vertices and removed. In this study, we set $t$ to 0.6.

$$L_{cla} = \sum_{i=1}^{N} -\log c_{\sigma(i)} \tag{6}$$

The contour vertex reduction operation does not strictly distinguish a building corner vertex, but assigns a unique corresponding predicted vertex as a positive example for each corner vertex of the ground-truth polygon. Therefore, even if the extracted building contours are of poor quality, their shapes will not be destroyed.

The contours after reduction may still have one or more vertices close to one of the building's corner vertices in a few cases. Although leaving them alone would not reduce the overall accuracy of the proposed building extraction method, we would rather apply two low-cost strategies for a better visual effect. First, a simple non-maximum suppression (NMS) algorithm is applied.

Specifically, we sort the vertices after reduction by their confidence scores, and retain the vertex with the highest confidence score and remove the other vertices $z$ pixels (e.g., half of the average line segment length of the predicted contour) nearer to this vertex. We repeat this step until all the points are processed. Second, we remove those reductant vertices on building edges with angles greater than $8\pi/9$, based on the angle between adjacent vertices.

### 2.3.3 Implementation of the contour evolution module

Advanced convolutional neuron structures can be applied for the feature representation of vertices in the contour evolution module. However, it should be noted that the contours of the building are circular, and it is necessary to create a periodic structure by copying and connecting the features of the starting vertices to the end of the last vertex, and vice versa. The detailed schematic of the contour evolution module we designed is shown in Fig. 6. We first up-dimension the feature vector from 66 to 128 by a 1D convolution module. The feature vectors are then encoded using a detail-to-global information aggregation strategy. In the shallow layer of the structure, we encode the 128-D vertex features with the convolutional operation of a small kernel and gradually expand the size of the kernels as the network layer deepens, to expand the receptive field. We classify the encoding part into three sub-structures as detail information extraction, local information extraction, and global information extraction. Detail information extraction focuses on the current vertex features, which are the most important part. The local information extraction part enriches the network information by aggregating the features of neighboring vertices. Global information can provide an overall picture of the current building instance and prevents vertices from attaching to the wrong building instance, i.e., an adjacent building. A shortcut connection [62] operation is performed between each convolution in the encoding part.

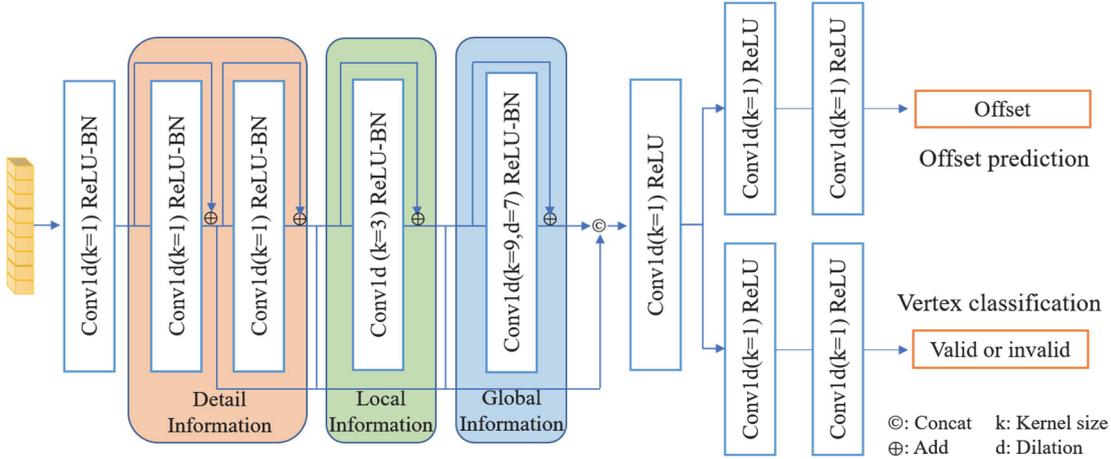

Fig. 6. Structure and parameters of the contour evolution module.

### 2.3.4 Loss function for contour supervision

Previous studies have paired the $N$ predicted vertices and ground-truth vertices statically, which can cause convergence difficulty or even wrong predictions. The dynamic matching loss (DML) we propose can handle this problem as it dynamically adjusts the relationship of the vertex pairing to supervise the contour prediction of the last contour evolution. The loss consists of two calculations: 1) the distance between the predicted vertex points (yellow) and the nearest point on the label boundary, which is represented by 10 times densified points (purple) for each segment, denoted as the black arrow in Fig. 7(a); and 2) the distance between each corner vertex (red) and its nearest predicted vertex, as shown in Fig. 7(b).

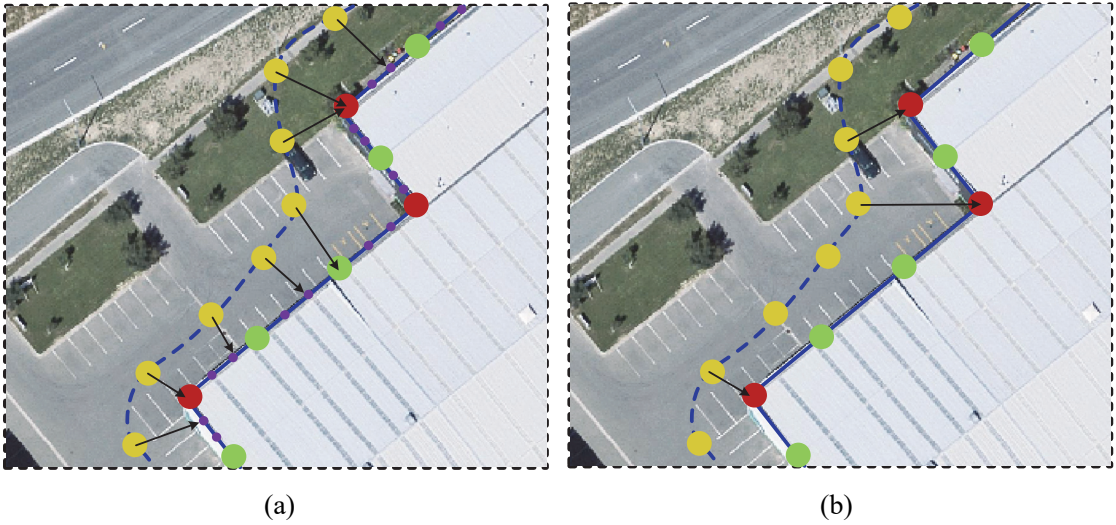

(a)                  (b)

Fig. 7. Example of dynamic matching loss. (a) A predicted contour vertex is pulled toward the nearest point on the densified label boundary. (b) A building corner vertex pulls the nearest predicted vertex toward its location. Yellow: predicted contour vertices. Red: corner vertices of the ground-truth polygon. Green: vertices on the edge of the ground-truth polygon. Purple: vertices of the 10 × densified ground truth $gt$.

In the first part, the interpolated ground truth $gt$ is denoted as $gt^{ipt} = \{p_j^{ipt}\}_{j=1}^{N \times 10}$. Eq. (7) formulates finding the nearest interpolated ground-truth contour vertex $x$ for the $i$-th point in the predicted vertices. The corresponding loss can be calculated by Eq. (8). The second loss of DML is concerned with improving the quality of the corner vertices. In the contour vertex reduction, we obtain the best predicted vertex with index $\sigma(i)$ for the $i$-th corner vertex of the current building $p^c$, so the second part can be calculated by Eq. (9). The DML is then the sum of Eqs. (8) and (9).

$$x_i^* = \arg\min \left\| \hat{p}_i - p_x^{ipt} \right\|_2 \tag{7}$$

$$L_{pre2gt}(pre, gt) = \frac{1}{N} \sum_{i=1}^{N} \left\| \hat{p}_i - p_{x_i^*}^{ipt} \right\|_1 \tag{8}$$

$$L_{gt2pre}(pre, gt) = \frac{1}{M} \sum_{i=1}^{M} \left\| \hat{p}_{\sigma(i)} - p_i^c \right\|_1 \tag{9}$$

Please note that DML is a fine-level adjustment loss that solves the over-smoothing problem existing in previous studies [26-28], so we use smooth $L_1$ loss to calculate the loss $L_{e1}$ for the first evolution, and DML to calculate the loss $L_{e2}$ for the second evolution.

## 3 Experimental settings

### 3.1 Datasets

In order to verify the performance and robustness of BuildMapper under large-scale and different datasets, three remote sensing building datasets, namely, the WHU aerial building dataset [63], the Crowd AI mapping challenge dataset [50], and the WHU-Mix (vector) building extraction dataset we built, were used to evaluate the proposed and compared methods.

### 3.1.1 WHU aerial building dataset

The WHU aerial building dataset [63] features a large quantity of high-resolution aerial images and corresponding highly accurate building labels. The aerial dataset covers 200 km² of the city of Christchurch, New Zealand, at a resolution of 0.2 m, and includes over 187,000 buildings of various architectures and purposes. All of the aerial images are cropped into 1024×1024 tiles. We transformed the corresponding building annotation label format from ".*shp*" to MS-COCO-style ".*json*" files [64]. The training set, validation set, and test set consist of 2793 (130,000 buildings), 627 (14,500 buildings), and 2220 (42,000 buildings) tiles, respectively.

**3.1.2 WHU-Mix (vector) building extraction dataset**

In another recent work [52], we described the large-scale and diverse building dataset named the WHU-Mix dataset, which was constructed through both editing the existing datasets and the addition of new samples. The WHU-Mix dataset consists of 64k image tiles with over 754k buildings, and covers an area of about 1100 km$^2$. Nevertheless, the WHU-Mix dataset is mainly intended for segmentation-based methods; however, in this work, the focus is on direct vector format building polygon extraction. To adapt to this, we edited the dataset and named it the WHU-Mix (vector) dataset. The WHU-Mix (vector) dataset uses the MS-COCO [64] format for the building labels.

Table 1 lists the composition details of the WHU-Mix (vector) dataset. The WHU-Mix (vector) dataset contains data from WHU [63], Crowd AI [50], Open AI [65], SpaceNet [49], and Inria [51] datasets, as well as five new data sets we collected. The annotation formats (GeoJson or Json) provided by Crowd AI, Open AI, and SpaceNet can be losslessly converted to the MS-COCO format. It should be noted that many label errors (dislocation, annotation missing, etc.) in the SpaceNet and Inria datasets have been manually corrected by us to guarantee the quality of WHU-Mix. New data sets are annotated in the shapefile format and can be easily converted into the MS-COCO format.

To validate the performance of the proposed method from multiple perspectives, we divided the entire dataset into a training set, a validation set, and two test sets (test sets I and II). Firstly, we mixed the Nos. 1–8 tiles and divided them into a training set (43,778 tiles), a validation set (2,922 tiles), and test set I (11,675 tiles), according to the ratio of 15:1:4. The Nos. 9–12 tiles were then mixed as test set II (6,011 tiles). The more challenging test set II does not have any geographical overlap with the training set, and can be used to further validate the generalization ability of a method on a more realistic real-world case.

**Table 1. The details of the WHU-Mix (vector) building extraction dataset**

| No. | Data source | Region | Resolution (m) | Tile size (pixel) | Tile quantity | Building quantity |
|---|---|---|---|---|---|---|
| 1 | Crowd AI [50] | - | 0.20 | 300 | 15,000 | 129,104 |
| 2 | SpaceNet [49]* | Vegas, Khartoum | 0.30 | 650 | 4,521 | 132,491 |
| 3 | Open AI [65] | Tanzania | 0.08 | 512 | 7,246 | 23,096 |
| 4 | WHU [63] | Christchurch | 0.20 | 512 | 14,531 | 237,674 |
| 5 | WHU | East Asia | 0.34 | 512 | 5,618 | 40,792 |
| 6 | New | Zhejiang | 0.20 | 512 | 6,009 | 40,183 |
| 7 | New | Hubei | 0.64 | 512 | 3,025 | 66,445 |
| 8 | New | Chongqing | 0.15 | 512 | 2,426 | 5,986 |
| 9 | Inria [51]* | Tyrol, Kitsap | 0.30 | 512 | 2,971 | 35,311 |
| 10 | SpaceNet* | Paris | 0.30 | 650 | 633 | 16,121 |
| 11 | New | Wixi | 0.17 | 512 | 1,267 | 6,305 |
| 12 | New | Dunedin | 0.20 | 512 | 1,140 | 20,618 |
| | | Total Quantity | | | 64,387 | 754,126 |

Note: '*' denotes that we edited the incorrect labels in these datasets.

**3.1.3 Crowd AI mapping challenge dataset**

The Crowd AI dataset [50] was treated as a secondary experimental dataset because it requires more computational cost with the large number of samples, but the samples lack diversity, which means that it performs much worse than the WHU-Mix (vector) dataset in properly evaluating a model's performance and generalization ability [52]. Nevertheless, as there are several building extraction studies that have reported their results on this dataset, we also report our results.

The Crowd AI dataset provides over 340k individual satellite imagery tiles of 300×300 pixel RGB images and building annotations. The training set and test set consist of 280,741 and 60,317 tiles, respectively, along with their corresponding annotations in MS-COCO format.

### 3.2 Evaluation metrics

We evaluated and compared the prediction results of the different methods from several perspectives using different evaluation metrics.

**Mask-based segmentation measures.** In this paper, the mask quality is evaluated in terms of the standard COCO measure of average precision (AP) under different mask intersection over union (IoU) thresholds. Mask IoU is defined as the ratio between the intersection and the union of the predicted contour *pre* and ground-truth contour *gt* (see Eq. (10)). Mask AP is an average of precision (i.e., the ratio of true positive and the sum of true positive and false positive) over 10 mask IoU values with thresholds from 0.50 to 0.95, with steps of 0.05. Thus, $AP^{msk}$ can be calculated as shown in Eq. (11).

$$\text{Mask IoU (G, P)} = \frac{pre \cap gt}{pre \cup gt} \tag{10}$$

$$AP^{mask} = \frac{AP_{0.50} + AP_{0.55} + \cdots + AP_{0.95}}{10} \tag{11}$$

**Boundary-based segmentation measures.** The number of internal pixels grows quadratically as the building size increases, and can far exceed the number of boundary pixels that only grow linearly. Mask-based segmentation measures values of all pixels equally and are less sensitive to prediction errors near building boundaries, especially in a larger building. Therefore, we use $AP^{bdy}$ [66] to further evaluate the quality of the building boundaries. Similar to $AP^{msk}$, $AP^{bdy}$ is calculated over multiple boundary IoU values. Given the predicted contour *pre* and ground-truth contour *gt*, the regions within the distance *d* from the boundary are calculated, and then the IoU of the two regions is calculated to obtain the boundary IoU. The boundary IoU is computed as follows:

$$\text{Boundary IoU}(gt, pre) = \frac{|(gt_d \cap gt) \cap (pre_d \cap pre)|}{|(gt_d \cap gt) \cup (pre_d \cap pre)|} \tag{12}$$

$AP^{bdy}$ can also be calculated as Eq. (11). In all the experiments, we set *d* to 1% of the image diagonal, for a fair comparison.

In addition, $AP_{S\&M}$ and $AP_L$ were used to further measure the performance of buildings of different sizes. *S&M* denotes small- and medium-sized buildings, and *L* denotes large buildings. In this paper, we consider buildings with less than 7500 pixels (about 300 m$^2$ on a 0.2-m resolution image) as being small- and medium-sized buildings, and the others as large buildings.

**Instance-based measures.** Replacing manual labor is the ultimate goal of automated building contour mapping. Following Wei et al. [22] we determined whether each predicted building instance contour satisfied the manual delineation level by simulating the errors that can be introduced by human operators when performing building contour annotation. We assumed a 2-pixel or 3-pixel manual delineation error introduced for delineating each edge, creating an instance-level IoU threshold for each building. Specifically, we extended the ground-truth polygon by 2 or 3 pixels and calculated the IoU with the original polygon as the instance-level threshold. For each building instance, we considered that the predicted contour reached the manually delineated level if the IoU between the predicted contour and the ground-truth contour was greater than the threshold corresponding to that instance.

### 3.3 Implementation details

The proposed BuildMapper optimizes multiple task losses jointly, including the building center point prediction loss $L_{ct}$, the initial contour prediction loss $L_{init}$, the first contour evolution loss $L_{py1}$, the second contour evolution loss $L_{py2}$, and the vertex classification loss $L_{cla}$. Therefore, we define a multi-task loss function:

$$L = L_{ct} + \varepsilon(L_{init} + L_{e1} + L_{e2}) + L_{cla} \quad (13)$$

where $\varepsilon$ is empirically set to 1/3. We implemented and tested BuildMapper in PyTorch on a desktop computer with an Nvidia RTX 3090 24 GB graphics processing unit (GPU). In the network training stage, we used multi-scale scaling, flipping, cropping, and color jittering as data augmentation, and used the Adam optimizer [67] to optimize the entire network.

Table 2 lists the assigned values and recommended value ranges of key parameters in BuildMapper. BuildMapper predicts a maximum number of K buildings in a single image, and a building contour with corner vertices less than N, both of N and K can be set to a relatively large value. T and $t$ are used to filter out invalid building instances and contour vertices, respectively. The iteration number *iter* is set to 2 as more iterations of the contour evolution module do not bring accuracy improvement. The distance cost weight $\delta$ in the Hungarian algorithm is set to 5. Although these parameters should be preset empirically, their fixed values (the third column in Table 2) are experimentally proved applicable on all the three datasets, that is to say, such parameter settings are very robust and consistent on various aerial images (WHU), satellite images (Crowd AI) and their combination (WHU-Mix).

**Table 2. Key parameter settings in BuildMapper.**

| Key parameters | Description | value | Recommended range |
|---|---|---|---|
| N | Number of building contour vertices | 64 | 64-128 |
| T | Threshold for valid building selection | 0.2 | 0.05-0.2 |
| K | Maximum number of buildings in a single image | 200 | 200-300 |
| *t* | Threshold for valid contour vertices selection | 0.6 | 0.5-0.7 |
| *iter* | Iterations number of contour evolution module | 2 | 2-3 |
| $\delta$ | Distance cost weight in the Hungarian algorithm | 5 | 4-6 |

We used different training schedules for the WHU aerial dataset, the WHU-Mix (vector) dataset, and the Crowd AI dataset (see Table 3), due to the large differences in the number of samples. The learning rates (*lr*) were all initially set to 1e-4, and decreased by 5× at $E_{d1}$ and $E_{d2}$ epochs. We first trained the contour initialization module for $E_{int}$ epochs, and then trained the whole network for $E_{total}$ epochs.

**Table 3. Training schedule for BuildMapper on different datasets**

| Dataset | First *lr* decay ($E_{d1}$) | Second *lr* decay ($E_{d2}$) | Initial training epoch ($E_{int}$) | Total epochs ($E_{total}$) |
|---|---|---|---|---|
| WHU aerial | 65 | 80 | 10 | 100 |
| WHU-Mix (vector) | 20 | 24 | 2 | 30 |
| Crowd AI | 16 | 20 | 2 | 26 |

## 4 Results and discussion

### 4.1 Comparison with the state-of-the-art methods

In this section, we describe how we first compared BuildMapper with the advanced mask-based and contour-based general instance segmentation methods, as well as the recent building extraction methods, on the WHU aerial building dataset and WHU-Mix (vector) dataset. We then describe how we compared the specified building extraction methods on the Crowd AI dataset. It should be clarified that, in these experiments, we only evaluated the performance in general instance segmentation (i.e., on polygons without vertex reduction).

Tables 4–6 list the quantitative comparison of BuildMapper, Mask R-CNN [17], SOLO [18], YOLACT [19], PolarMask [40], Deep Snake [27], and CLP-CNN [29] for mask AP, boundary AP, and speed (frames per second, FPS) on the WHU aerial building dataset and WHU-Mix (vector) dataset. For a fair comparison, we report the results of the proposed method for both DLA-34 and ResNet-50 with FPN (R-50-FPN) backbones.

For the WHU building dataset (Table 4), BuildMapper achieves the best accuracy at the highest inference speed, with a score of 78.5 $AP^{msk}$ and 72.8 $AP^{bdy}$ at 22.4 FPS, which outperforms the second-best CLP-CNN method by 3.9% in $AP^{msk}$ and 4.6% in $AP^{bdy}$ and the classic Mask R-CNN 14.3% in $AP^{msk}$ and 14.8% in $AP^{bdy}$. The CLP-CNN method uses a concentric loop convolutional structure for building contour adjustment, which is advantageous for the extraction of global information about buildings. However, this structure also reduces the sensitivity of the network to detailed information. At the same time, the concentric loop structure requires more computation, and its inference is correspondingly slower (about half as fast as BuildMapper). The performance of BuildMapper with R-50-FPN or DLA-34 as the backbone is close. Using DLA-34 as the backbone can bring some accuracy improvement, but it also leads to a slight decrease in speed. In this comprehensive comparison, the contour-based methods show a significant performance advantage over the mask-based methods on the high-resolution WHU dataset. This suggests that it is more reasonable to directly extract building contours by vertex regression than by pixel-level classification.

Table 4. Quantitative comparison on the WHU aerial building dataset.

| Method | Backbone | FPS | $AP^{msk}$ | $AP^{msk}_{S\&M}$ | $AP^{msk}_L$ | $AP^{bdy}$ | $AP^{bdy}_{S\&M}$ | $AP^{bdy}_L$ |
|---|---|---|---|---|---|---|---|---|
| *Mask-based:* | | | | | | | | |
| Mask R-CNN | R-50-FPN | 7.0 | 63.3 | 70.2 | 45.7 | 57.4 | 66.0 | 29.4 |
| YOLACT | R-50-FPN | 12.2 | 58.5 | 61.7 | 41.1 | 52.5 | 57.5 | 27.7 |
| SOLO | R-50-FPN | 6.1 | 67.9 | 71.4 | 56.7 | 59.9 | 66.3 | 38.2 |
| *Contour-based:* | | | | | | | | |
| PolarMask | R-50-FPN | 5.3 | 61.3 | 64.9 | 41.4 | 51.0 | 57.7 | 15.2 |
| Deep Snake | Dla-34 | 17.6 | 72.2 | 75.4 | 48.8 | 66.3 | 71.4 | 34.0 |
| CLP-CNN | Dla-34 | 11.2 | 74.6 | 78.0 | 58.8 | 68.2 | 73.9 | 41.2 |
| BuildMapper | R-50-FPN | 24.1 | 77.6 | 81.4 | 48.5 | 72.2 | 77.7 | 37.2 |
| BuildMapper | Dla-34 | 22.4 | **78.5** | **82.1** | **59.5** | **72.8** | **78.6** | **44.1** |

For the WHU-Mix (vector) test set I, BuildMapper also outperforms the other methods in all metrics, as shown in Table 5, which is consistent with the results obtained on the WHU aerial building dataset. The $AP^{msk}$ score of the recent SOLO method approaches that of the top contour-based methods, but its $AP^{bdy}$ score and speed are significantly lower. We can find that using DLA as the backbone in BuildMapper performs slightly better on the WHU-Mix (vector) dataset (Table 5&6). In contrast to the WHU dataset, the WHU-Mix (vector) dataset incorporates a variety of data of different imaging qualities and building styles from multiple sources. Therefore, we believe that the DLA backbone is a better choice as a lightweight backbone for the contour-based methods.

Table 5. Quantitative comparison on the WHU-Mix (vector) test set I.

| Method | Backbone | FPS | $AP^{msk}$ | $AP^{msk}_{S\&M}$ | $AP^{msk}_L$ | $AP^{bdy}$ | $AP^{bdy}_{S\&M}$ | $AP^{bdy}_L$ |
|---|---|---|---|---|---|---|---|---|
| *Mask-based:* | | | | | | | | |
| Mask R-CNN | R-50-FPN | 38.9 | 47.0 | 45.8 | 62.8 | 24.7 | 26.8 | 8.0 |
| YOLACT | R-50-FPN | 40.1 | 42.3 | 41.3 | 55.7 | 20.3 | 21.6 | 6.4 |
| SOLO | R-50-FPN | 24.8 | 57.1 | 56.2 | 69.4 | 23.9 | 25.4 | 10.2 |
| *Contour-based:* | | | | | | | | |
| PolarMask | R-50-FPN | 15.7 | 44.8 | 43.7 | 59.5 | 17.0 | 18.9 | 2.2 |
| Deep Snake | Dla-34 | 41.6 | 55.3 | 54.9 | 62.8 | 23.6 | 25.4 | 6.5 |
| CLP-CNN | Dla-34 | 33.2 | 59.2 | 58.7 | 66.9 | 28.7 | 30.4 | 10.9 |
| **BuildMapper** | R-50-FPN | 61.8 | 58.3 | 58.0 | 65.0 | 28.7 | 30.1 | 11.5 |
| **BuildMapper** | Dla-34 | 64.5 | **59.9** | **59.4** | **68.6** | **29.9** | **31.4** | **12.9** |

Test set II of the WHU-Mix (vector) dataset is the most challenging because it does not have any regions that are geographically adjacent to the regions of the training set. BuildMapper shows a stable and excellent performance on test set II. Furthermore, BuildMapper outperforms all the other methods in the stringent boundary-based metrics. BuildMapper and CLP-CNN models obtain similar high accuracies in the mask-based metrics, but the latter requires far more parametric and computational effort than BuildMapper, as mentioned above. More details of the parametric and computational cost are provided in Section 5.3.4. For the segmentation-based methods, Mask R-CNN slightly outperforms the most recent SOLO method, showing its robustness to different situations by comparing Tables 5 and 6. We did not list the inference speed because it is almost the same as in Table 5.

Table 6. Quantitative comparison on the WHU-Mix (vector) test set II.

| Method | Backbone | $AP^{msk}$ | $AP^{msk}_{S\&M}$ | $AP^{msk}_L$ | $AP^{bdy}$ | $AP^{bdy}_{S\&M}$ | $AP^{bdy}_L$ |
|---|---|---|---|---|---|---|---|
| *Mask-based:* | | | | | | | |
| Mask R-CNN | R-50-FPN | 46.1 | 45.9 | 49.2 | 23.1 | 25.2 | 4.3 |
| YOLACT | R-50-FPN | 41.3 | 40.7 | 49.6 | 18.4 | 19.9 | 3.9 |
| SOLO | R-50-FPN | 45.3 | 44.6 | 55.6 | 19.7 | 21.6 | 4.7 |
| *Contour-based:* | | | | | | | |
| PolarMask | R-50-FPN | 39.1 | 38.6 | 45.2 | 15.2 | 16.5 | 1.3 |
| Deep Snake | Dla-34 | 46.9 | 46.5 | 54.6 | 22.7 | 24.6 | 4.4 |
| CLP-CNN | Dla-34 | **49.7** | **49.1** | 58.4 | 27.9 | 29.9 | 8.2 |
| **BuildMapper** | R-50-FPN | 48.1 | 47.6 | 55.0 | 27.4 | 29.0 | 8.2 |
| **BuildMapper** | Dla-34 | **49.7** | **49.1** | 59.0 | 28.5 | 30.3 | 9.6 |

According to Table 4–6, the $AP^{bdy}$ scores of small and medium buildings are significantly better than that of large buildings, which is the natural results of the calculation method of $AP^{bdy}$. For $AP^{msk}$, the difference is not that significant. In the WHU-Mix (vector) test set, $AP^{msk}$ of smaller buildings is better than that of larger ones; while in the WHU test set, the result is just the opposite. This is caused by the difference of the two datasets where the latter has larger image size, which may form a better condition for detecting larger buildings.

We also report the results of the proposed method obtained on the Crowd AI test set in Table 7, and compare it with the results of the most recent building extraction methods from the corresponding papers (which provide only the mask-based segmentation measures). BuildMapper shows a state-of-the-art performance. For the 300×300 images in the Crowd AI dataset, the proposed algorithm runs at 71.2 FPS. The HTC-DP [23] and frame field learning (FFL) [16] methods are both mask-based methods, and the latter achieves the second-best result. The building outline delineation (BOD) [37] and PolyMapper [38] methods use an RNN to iteratively predict the building corner vertices, which is prone to missing vertices and leads to a relatively poor accuracy.

Table 7. Quantitative comparison on the Crowd AI dataset.

| Method | $AP^{msk}$ | $AP_{50}^{msk}$ | $AP_{75}^{mask}$ |
|---|---|---|---|
| HTC-DP | 44.4 | - | - |
| BOD | 47.4 | - | - |
| PolyMapper | 55.7 | 86.0 | 65.1 |
| FFL | 61.3 | 87.4 | 70.6 |
| **BuildMapper** | **63.9** | **90.1** | **75.0** |

**4.2 Manual level of building annotation**

Removing the redundant vertices of the polygons yields the final prediction of BuildMapper. In this way, we can count how many automatically delineated buildings reach the level of manual annotation. Table 8 lists the percentage of buildings that reach the manual delineation level under different threshold conditions. We can see that 84.2% of the buildings can reach the manual delineation level at a 2-pixel accuracy on the high-quality WHU dataset. If the restriction is loosened to the 3-pixel level, 89.9% of the buildings can reach the manual delineation level. We also list the very few studies that have reported the results of structured vector format building extraction. The method proposed in an earlier study [22], which consists of a semantic segmentation network and empirical regularization post-processing, reported that 63% of the building contours can reach a manual delineation level under a 3-pixel accuracy. The method proposed in the latest study [29], which consists of a contour-based network and regularization post-processing, reported that 85.2% of the building contours can reach a manual delineation level. In fact, from previous studies [22, 29] and this work, we have found that a 3-pixel accuracy is sufficient to measure the manual delineation level. This means that the vast majority of the automatically delineated buildings, as high as 90%, can be used directly for practical applications such as map updating. It should be noted that, in [22], a fixed threshold was used to discriminate whether a building contour reached the manual delineation level, and thus the value of 63% may be slightly underestimated; however, it still cannot reach the level of the most advanced approaches.

In the Crowd AI dataset, which has a poorer data quality, 68.3/85.2% of the building contours obtained by the proposed method meet the 2-pixel/3-pixel criterion. Nevertheless, both the WHU and Crowd AI datasets were collected from a certain region with similar building structures and image appearances. In the WHU-Mix (vector) test set I, the complex and diverse building types come from global cites in five continents, which causes the drop in the automation level, compared to the Crowd AI and WHU aerial datasets. In the WHU-Mix (vector) test set II, the drop is more obvious; however, test set II simulates a real-world case, where the model was pretrained on a large and well-prepared dataset, and the target images are from other cities never seen by the model before. Therefore, the results obtained for the WHU-Mix (vector) test set II are very encouraging, and in this challenging real-world case, more than 60% of the building contours can be produced automatically, to replace human work.

Table 8. The percentage of extracted buildings that reach the manual delineation level with BuildMapper.

| Dataset | | 2-pixel level | 3-pixel level |
|---|---|---|---|
| WHU | [22] | 34% | 63% |
| | [29] | 77.4% | 85.2% |
| | Proposed | 84.2% | 89.9% |
| Crowd AI | | 68.3% | 85.2% |
| WHU-Mix (vector) test set I | | 60.0% | 74.6% |
| WHU-Mix (vector) test set II | | 50.8% | 62.5% |

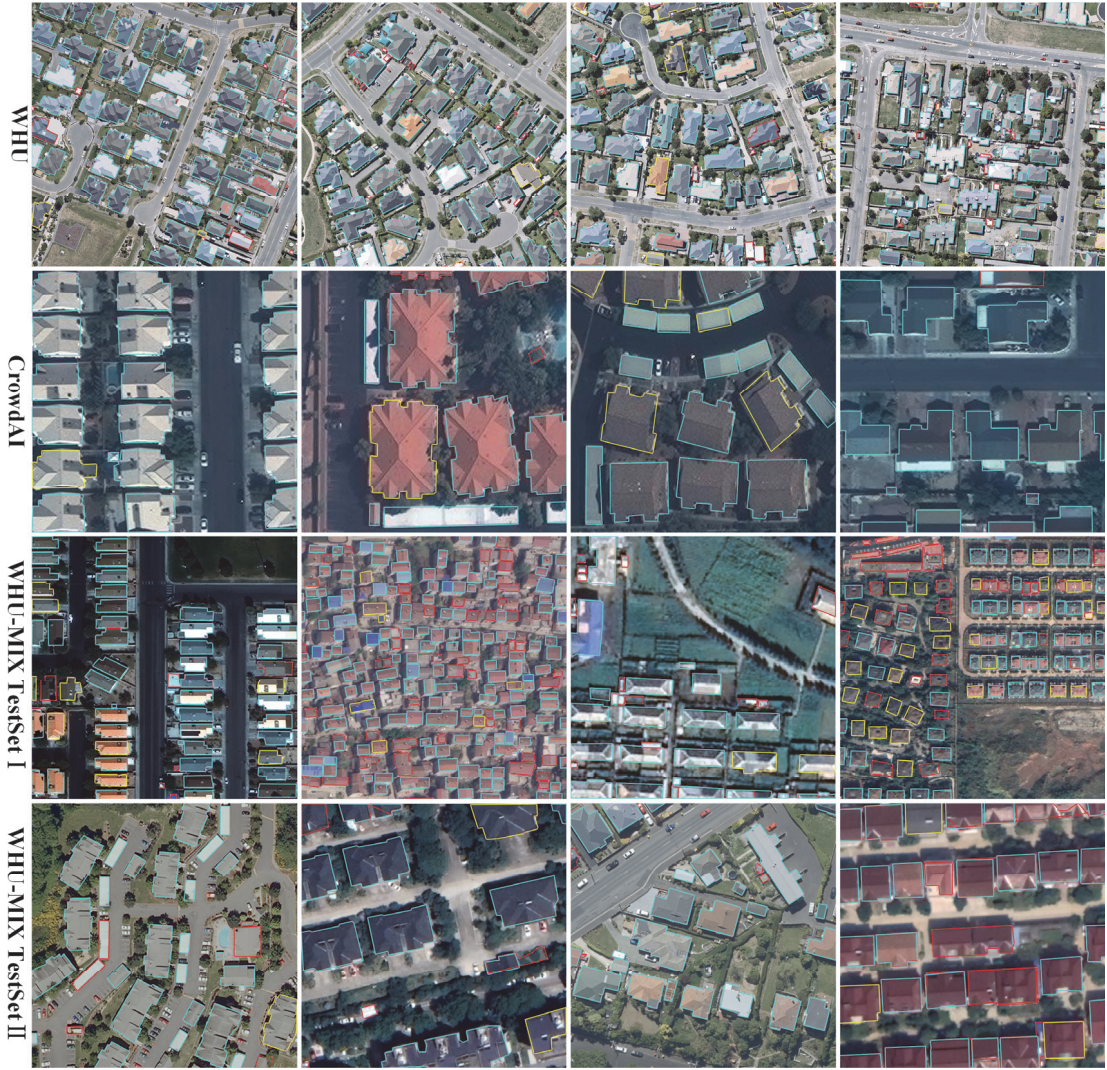

Fig. 8. Result of BuildMapper obtained on the different datasets. Red polygons: have not reached the manual delineation level; yellow polygons: reached the 3-pixel accuracy level; blue polygons: reached the 2-pixel accuracy level.

Fig. 8 shows the final output of BuildMapper, where the red polygons are those that have not reached the manual delineation level, the yellow polygons are those that have reached the 3-pixel accuracy level, and the blue polygons are those which have reached the 2-pixel accuracy level. We see the 3-pixel accuracy is sufficient because all of the yellow polygons have shown reached manual delineation level. The red polygons in the WHU dataset are mainly small artificial objects that are not labeled in the ground truth, which can be easily removed based on area. There are also many red building polygons, especially in test sets I and II, that are correct, which further confirms the rigor and correctness of the threshold setting of the manual delineation level. The Crowd AI dataset has poorer-quality images and the contour details extracted by BuildMapper are slightly worse than the results obtained on the WHU dataset. The WHU-Mix (vector) dataset shows the results for different types of buildings in different regions around the world. The second column of the third row of Fig. 8 is a dense area of small buildings, and the building instances in this area are difficult to label, even manually. However, the proposed method can extract most of the buildings correctly and reaches the manual delineation level. This is almost infeasible with the semantic segmentation based methods. Although the building styles in the three datasets differ significantly, especially in the WHU-Mix (vector) dataset, most of the buildings can be extracted accurately. This indicates that BuildMapper has great potential for automated building contour production.

Fig. 9 further shows the extraction results for different styles and architectures of individual buildings in the different datasets. Whether it is a tall building, a bungalow, a small residential house, or a large factory, the proposed method can extract its contour stably and correctly. It can be noted that some images, e.g., the second row and fourth column of Fig. 9, are non-orthogonal, and the rectangular buildings show a diamond shape in the image; it is difficult to generalize their contours by simple regularization rules based on empirical knowledge. For example, in the method proposed in [22], all the edges of a building should be adjusted parallel or perpendicular to the principal direction. Instead, BuildMapper can output regularized building contours consisting of corner vertices directly, eliminating the need for additional post-processing.

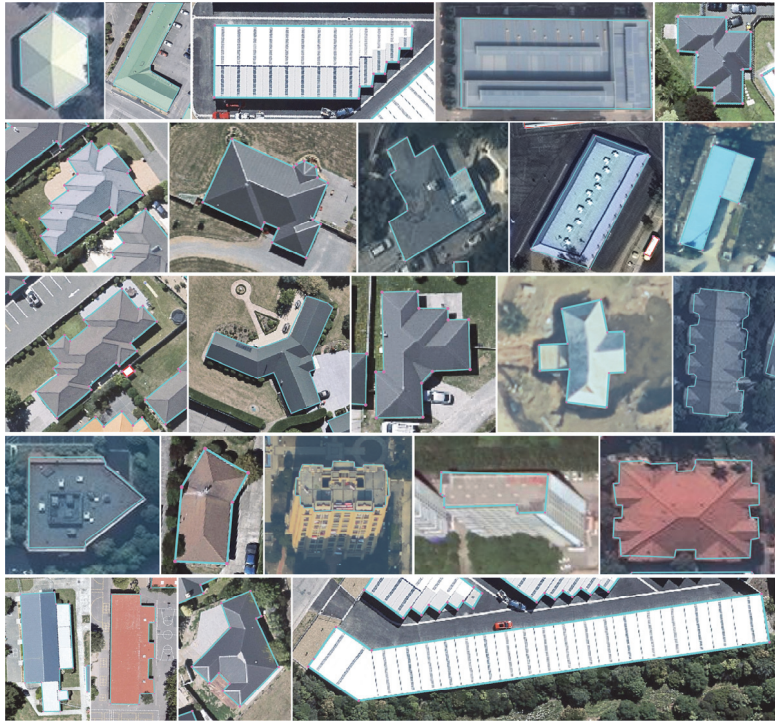

Fig. 9. Examples of different-style building contours generated by BuildMapper.

### 4.3 Ablation study

In the ablation study, we first verified the functions of the different stages of BuildMapper on all the datasets. We then evaluated the feasibility of BuildMapper by using alternative design solutions for each stage on the WHU dataset.

**4.3.1 Building contour accuracy in the different stages**

The proposed BuildMapper is a multi-stage approach. Table 9 lists the results of BuildMapper at the different stages on the three datasets. The initial contour accuracy of BuildMapper is already competitive with the other methods in Tables 4–7. The difference between the loose $AP_{50}$ indicator of the initial contour and the final result is small. However, it can be seen from Fig. 10 that the initial contours are smooth. Nevertheless, the learnable initial contour extraction method is much more accurate than those methods using empirically designed shapes (e.g., eclipse) as the initial contours, which is detailed in Section 4.3.2. After the first contour evolution based on the feature information of each vertex, the building contour quality is significantly improved, but the performance of the detailed part (especially the corner vertices) is still inferior. In the second evolution, we use DML to supervise the network, and the details of the contours are significantly improved. The vertex reduction of the contour evolution module can remove the vast majority of the redundant vertices. Consistent with the need for manual annotation, the final output of BuildMapper retains only the corner vertices of the buildings, and thus looks very regular, although the inclusion of the vertex reduction leads to a slight decrease in accuracy.

**Table 9. Results for the different stages of BuildMapper.**

| Dataset | Stage | $AP^{msk}$ | $AP^{msk}_{50}$ | $AP^{msk}_{75}$ | $AP^{bdy}$ | $AP^{bdy}_{50}$ | $AP^{bdy}_{75}$ |
|---|---|---|---|---|---|---|---|
| WHU | Initial contour | 71.6 | 93.2 | 85.2 | 60.8 | 91.7 | 73.1 |
| | 1st evolution | 77.7 | 93.3 | 87.4 | 71.4 | 93.1 | 83.7 |
| | 2nd evolution | 78.5 | 93.4 | 87.4 | 72.8 | 93.2 | 84.1 |
| | Vertex reduction | 77.0 | 93.3 | 86.4 | 70.7 | 92.2 | 82.1 |
| WHU-Mix (vector) test set I | Initial contour | 56.1 | 83.7 | 65.3 | 20.5 | 55.4 | 9.8 |
| | 1st evolution | 59.7 | 84.6 | 68.1 | 28.4 | 63.0 | 23.3 |
| | 2nd evolution | 60.0 | 84.6 | 68.1 | 29.9 | 64.2 | 25.3 |
| | Vertex reduction | 59.1 | 83.8 | 67.3 | 28.6 | 62.5 | 23.6 |
| WHU-Mix (vector) test set II | Initial contour | 46.9 | 73.6 | 51.8 | 21.2 | 55.6 | 11.3 |
| | 1st evolution | 49.6 | 74.4 | 54.5 | 27.2 | 60.7 | 21.8 |
| | 2nd evolution | 49.7 | 73.8 | 54.5 | 28.5 | 61.9 | 23.9 |
| | Vertex reduction | 48.8 | 73.0 | 53.2 | 27.3 | 60.2 | 22.3 |
| Crowd AI | Initial contour | 62.0 | 89.2 | 73.7 | 8.3 | 35.5 | 0.9 |
| | 1st evolution | 63.9 | 90.1 | 75.0 | 10.8 | 43.1 | 1.6 |
| | 2nd evolution | 63.9 | 90.1 | 75.0 | 11.4 | 44.5 | 1.8 |
| | Vertex reduction | 63.6 | 89.3 | 74.9 | 10.6 | 42.2 | 1.6 |

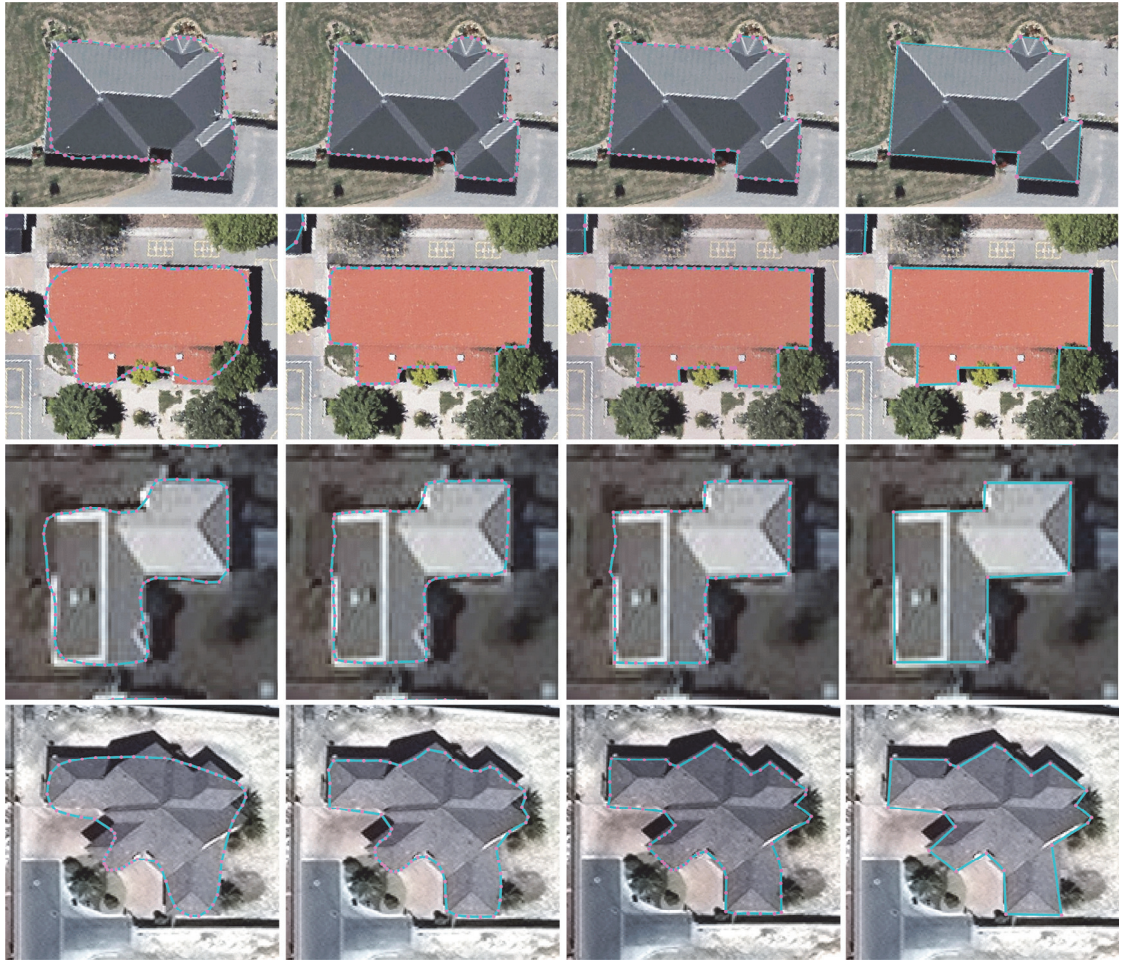

      (a) Initial contour          (b) 1st evolution          (c) 2nd evolution         (d) Vertex reduction

Fig. 10. Examples of building contour extraction results at the different stages of BuildMapper.

### 4.3.2 Different contour initialization methods

We examined the effect of different contour initialization methods for BuildMapper on the WHU dataset, including the ellipse of Curve GCN, the rectangle of DANCE, and the octagon of Deep Snake. Since these manually designed contour initialization methods rely on the bounding box of the building being detected, we had to add an additional bounding box detection branch. We first verified the impact of introducing such a branch into BuildMapper. For each building instance, the bounding box was obtained by predicting the width and height of the building based on the features of the center point. As can be seen in Table 10, adding the box branch leads to a slight decrease in accuracy. We speculate that this may be due to the ambiguity caused by the contour initialization branch and the box branch sharing the same features as the input. However, adding an additional box branch is acceptable.

Given the bounding box of a building, rectangles and ellipses can be easily obtained according to the position of the box. For the octagon of Deep Snake, we used the features at the midpoints of the four edges of the box to predict the four extreme points (top-most, left-most, bottom-most, right-most) of a building, and then composed the octagon based on the locations of the bounding box and the extreme points. Table 11 shows that the accuracies of the initial contours obtained by the other methods are much lower than the accuracy of the contours obtained by our learnable instance-aware contour initialization approach. Fig. 11 shows that the contour initialization methods with manually designed shapes obtain very inadequate contours. Meanwhile, the network architecture becomes more complicated due to the additional operations such as bounding box prediction and contour resampling. Although the other initial contours are of poor quality, the powerful contour evolution module in BuildMapper can still output satisfactory results from them.

Table 10. Effect of the box branch

|  | $AP^{msk}$ | $AP^{msk}_{50}$ | $AP^{msk}_{75}$ | $AP^{bdy}$ | $AP^{bdy}_{50}$ | $AP^{bdy}_{75}$ |
|---|---|---|---|---|---|---|
| With box branch | 78.3 | 93.3 | 87.3 | 72.4 | 93.1 | 84.0 |
| Without box branch | 78.5 | 93.4 | 87.4 | 72.8 | 93.2 | 84.1 |

Table 11. Comparison of the different contour initialization methods

| Initialization approach | Stage | $AP^{msk}$ | $AP^{msk}_{50}$ | $AP^{msk}_{75}$ | $AP^{bdy}$ | $AP^{bdy}_{50}$ | $AP^{bdy}_{75}$ |
|---|---|---|---|---|---|---|---|
| DANCE | Initial contour | 11.4 | 50.3 | 2.6 | 2.6 | 9.9 | 0.8 |
|  | 1st evolution (our) | 73.3 | 93.0 | 86.0 | 64.4 | 91.7 | 78.3 |
|  | 2nd evolution (our) | 77.6 | 93.1 | 87.1 | 71.8 | 92.8 | 83.2 |
| Curve GCN | Initial contour | 25.7 | 82.9 | 4.6 | 7.8 | 29.5 | 0.9 |
|  | 1st evolution (our) | 74.8 | 93.2 | 86.4 | 67.0 | 92.2 | 80.4 |
|  | 2nd evolution (our) | 77.7 | 93.3 | 86.7 | 72.3 | 92.5 | 83.6 |
| Deep Snake | Initial contour | 36.9 | 90.6 | 8.4 | 12.8 | 47.8 | 1.4 |
|  | 1st evolution (our) | 70.8 | 93.1 | 84.7 | 60.1 | 91.5 | 71.3 |
|  | 2nd evolution (our) | 77.6 | 93.2 | 86.5 | 71.5 | 93.0 | 83.0 |
| Proposed | Initial contour | 71.5 | 93.1 | 85.1 | 61.0 | 91.7 | 73.4 |
|  | 1st evolution | 77.3 | 93.2 | 87.2 | 71.1 | 93.0 | 82.9 |
|  | 2nd evolution | 78.5 | 93.4 | 87.4 | 72.8 | 93.2 | 84.1 |

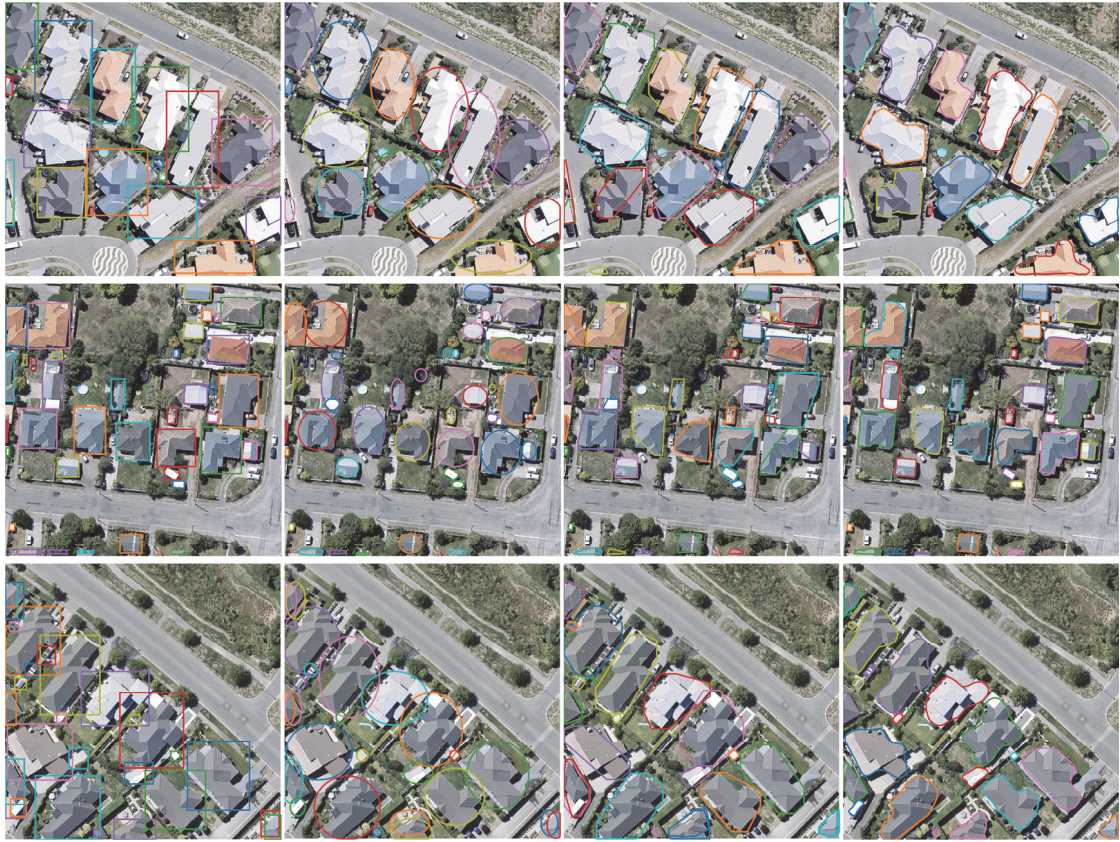

(a) DANCE      (b) Curve GCN      (c) Deep Snake      (d) BuildMapper

Fig. 11. The initial contours of the different methods. (a) The rectangle of DANCE; (b) the ellipse of Curve GCN; (c) the octagon of Deep Snake; (d) learnable initial contour of BuildMapper.

### 4.3.3 Loss function for supervision

We compare the results of the last evolution module being supervised by smooth $L_1$ loss, Chamfer loss, and DML in Table 12. The advantage of the proposed DML is very obvious against the other two losses, and it gains improvements of 1.2 $AP^{msk}$ and 1.7 $AP^{bdy}$ over the first evolution. Smooth $L_1$ loss and Chamfer loss perform similarly. Compared to the first evolution result, these two loss functions bring only a very slight improvement. Fig. 12 illustrates the quality of the building contours obtained under different loss functions. The overall visual effect of the building contour with DML as the loss function is significantly better, especially for the corner vertex region. The accurate inflection points also provide good conditions for the subsequent contour vertex reduction.

**Table 12. The results of the last evolution module supervised by different loss functions**

| Stage | Loss | $AP^{msk}$ | $AP^{msk}_{50}$ | $AP^{msk}_{75}$ | $AP^{bdy}$ | $AP^{bdy}_{50}$ | $AP^{bdy}_{75}$ |
|---|---|---|---|---|---|---|---|
| 1st evolution | -- | 77.3 | 93.2 | 87.2 | 71.1 | 93.0 | 82.9 |
| 2nd evolution | Smooth $L_1$ | 77.4 | 93.9 | 87.2 | 71.2 | 93.0 | 83.5 |
|  | Chamfer | 77.4 | 93.3 | 86.5 | 71.6 | 92.4 | 83.0 |
|  | DML | 78.5 | 93.4 | 87.4 | 72.8 | 93.2 | 84.1 |

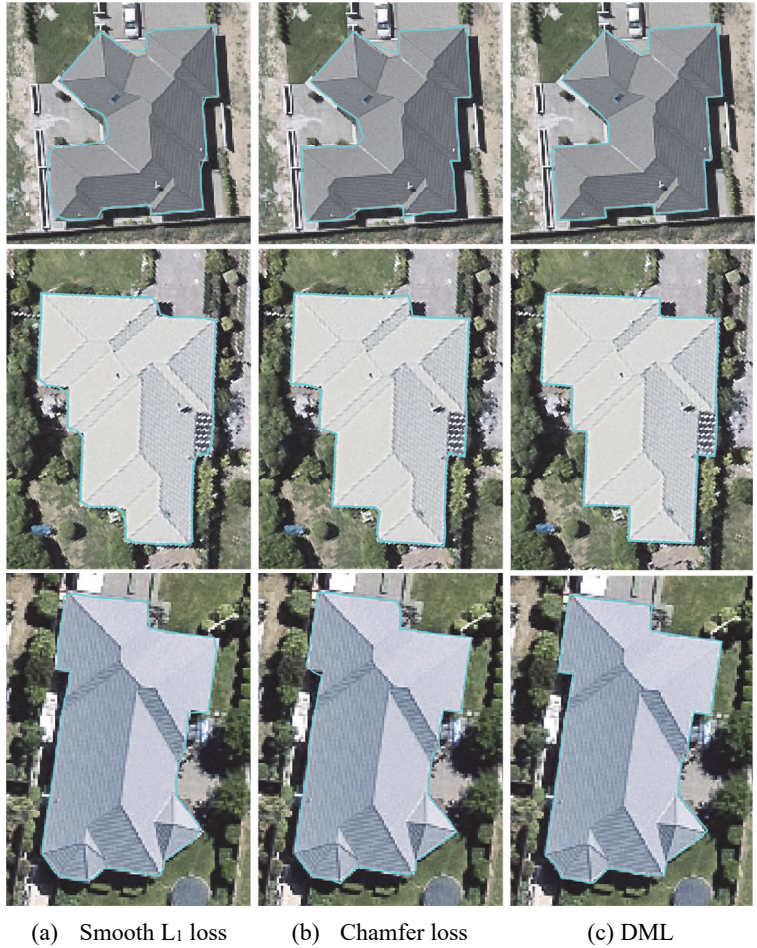

(a) Smooth $L_1$ loss    (b) Chamfer loss    (c) DML

Fig. 12. Contour results obtained with different loss functions.

### 4.3.4 Different contour evolution structures

Table 13 lists the evaluation results of the proposed method using different contour evolution modules. We replaced the contour evolution module of BuildMapper with the CIR-Conv module of Deep Snake and the CLP-Conv module of CLP-CNN, respectively, and evaluated their accuracy. The CIR-Conv module aggregates the features of the vertices in the neighborhood in a simple way, and its performance is relatively poor. CLP-Conv uses a concentric loop structure to aggregate the instance details and global information, and its performance is very close to that of the proposed method, but the computational effort (Flops) and the network parameters (Params) are much higher than those of the proposed method, as are those of CIR-Conv. The contour evolution module proposed in this paper adopts a detail-to-global information aggregation strategy, which can guarantee both accuracy and a very high efficiency.

Table 13. Comparison of the different contour evolution modules.

|  | Flops (G) | Params (M) | $AP^{msk}$ | $AP^{msk}_{50}$ | $AP^{msk}_{75}$ | $AP^{bdy}$ | $AP^{bdy}_{50}$ | $AP^{bdy}_{75}$ |
|---|---|---|---|---|---|---|---|---|
| CIR-Conv | 2.44 | 3.78 | 78.0 | 93.3 | 86.4 | 72.5 | 93.1 | 83.8 |
| CLP-Conv | 10.92 | 5.73 | 78.3 | 93.0 | 87.1 | **73.2** | 92.8 | **84.8** |
| Proposed | 0.35 | 0.64 | **78.5** | **93.4** | **87.4** | 72.8 | **93.2** | 84.1 |

### 4.3.5 Vertex reduction

One of the innovations of the proposed BuildMapper is integrating the vertex reduction into an end-to-end learnable framework, while other methods have to introduce additional post-processing operations onto the outputs of a deep learning model to obtain structured building contours. We compared the learnable vertex reduction module with the post regularization method in [22]. We removed the vertex reduction module and applied the empirical regularization method in [22] onto the outputs of BuildMapper. In Table 14, the proportion of manual delineation level buildings obtained by the learnable vertex reduction module is 1.3% higher than that obtained by the empirical post-processing regularization in the 2-pixel accuracy level, and is equivalent to it under the 3-pixel accuracy. This indicates that the learnable method can discover and delineate more details of the building boundary. Fig. 13 shows two examples. The empirical regularization algorithm, which is restricted by the empirical rule and fixed parameters, cannot handle the smoothing of a small hump (the first row) and the obtuse angles consisting of short edges (the second row), although such small output differences between the two methods would not impact the measurement of 3-pixel accuracy. In contrast, the learnable method performs much more flexibly in the various situations.

Table 14. Comparison between the learnable vertex reduction module and the empirical regularization method.

| Method | 2-pixel level | 3-pixel level |
| --- | --- | --- |
| Regularization [22] | 82.9% | 90.2% |
| Vertex reduction (proposed) | 84.2% | 89.9% |

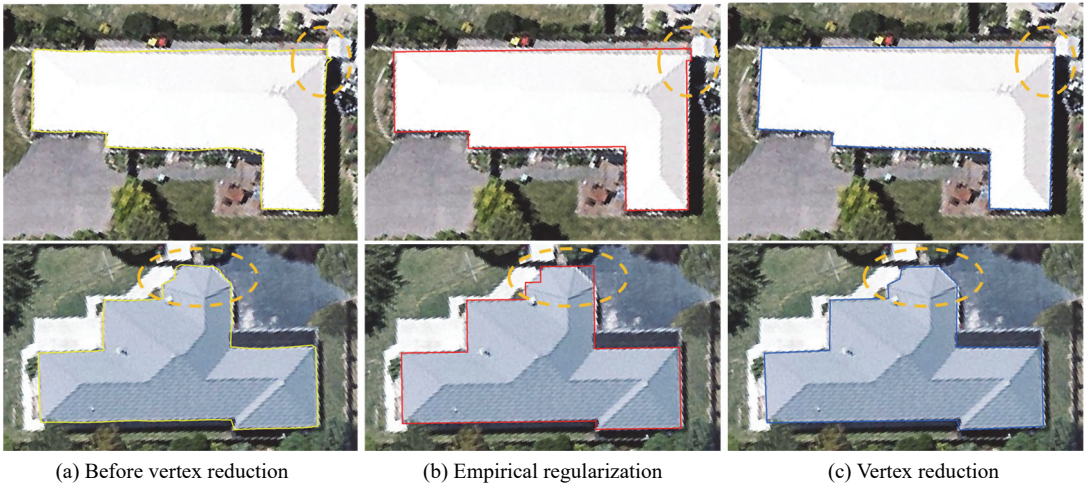

(a) Before vertex reduction　　　(b) Empirical regularization　　　(c) Vertex reduction

Fig. 13. Examples of empirical regularization and vertex reduction.

### 4.4 Generalization ability of different datasets

In practical applications, there can be significant differences of image sources, geographical regions, and building styles between a training set and target images. A diverse large-scale training dataset can greatly improve the generalization ability of a trained model. Here we evaluate the performance of BuildMapper respectively trained on the WHU-Mix (vector), WHU Aerial, and Crowd AI training datasets on the test set II of WHU-Mix (vector), as the test set II does not overlap with any training sets in terms of geographic regions. According to Table 1, the model trained on the WHU dataset have the worst generalization ability as it only consists of aerial images in Christchurch. The model trained on the Crowd AI dataset is slightly better, while the model trained on the WHU-Mix (vector) training set shows a much better performance than the rest. It demonstrates that a large-scale dataset containing images from different platforms, sensors, and regions can significantly improve the generalization ability of a building extraction model. This is one of the key reasons for us to create the WHU-Mix dataset.

**Table 15. Results of BuildMapper pretrained on different training datasets and applied to the WHU-Mix (vector) test set II.**

| Training dataset | $AP^{msk}$ | $AP^{msk}_{50}$ | $AP^{msk}_{75}$ | $AP^{bdy}$ | $AP^{bdy}_{50}$ | $AP^{bdy}_{75}$ |
|---|---|---|---|---|---|---|
| WHU Aerial | 23.0 | 31.6 | 25.6 | 14.6 | 28.6 | 14.3 |
| Crowd AI | 32.4 | 51.3 | 36.2 | 13.7 | 37.0 | 6.9 |
| WHU-Mix (vector) | 49.7 | 73.8 | 54.5 | 28.5 | 61.9 | 23.9 |

**4.5 Limitations and future work**

The excellent performance of BuildMapper has been demonstrated in this paper, and the newest and highest automation level has been reported, but BuildMapper does have the following limitations. Firstly, BuildMapper extracts the outermost contour of a building. If the building is a hollow object, as in Fig. 14(a), the background part in the center will not be eliminated. Secondly, BuildMapper locates building targets based on center points, which may cause false detection or missed detection. As shown in Fig. 14(b), the large cars are mistaken as buildings. In contrast, the individual buildings in Fig. 14(c) cannot be correctly identified as the center point of a building is missing due to cloud shading. Thirdly, if the predicted contours are of poor quality, the vertex reduction module may lead to some loss of details. Fig. 14(d) and (e) are the results before and after redundant vertex removal, respectively. The contour segments inside the yellow ellipses in (e) are slightly worse than those in (d). In addition, as shown in Fig. 14(f), the occlusion results in a missing building (the red building), although the proposed method can resist the impact of partial occlusion (the cyan building).

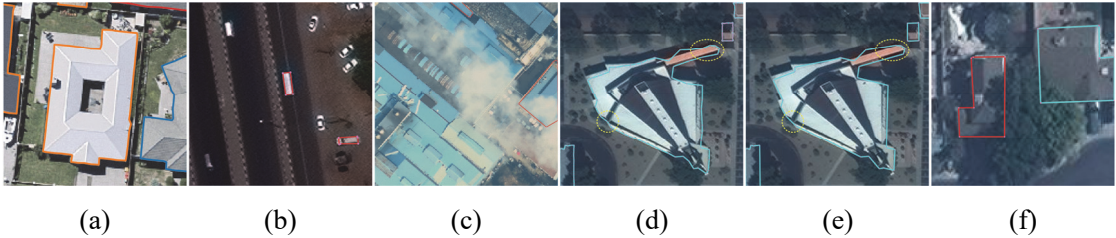

| (a) | (b) | (c) | (d) | (e) | (f) |

Fig. 14. Examples of buildings that are not correctly extracted by the proposed BuildMapper. (a) Building with a hollow; (b) cars mistaken as buildings; (c) buildings shaded by clouds; (d-e) incorrect redundant vertex removal; (f) buildings shaded by a tree.

In our future work, we will attempt to extract building contours by directly initializing corner vertices and their topological connection relationships, instead of by the center point. Compared to the center of a building, the corner vertices have more distinct textural characteristics and are therefore more stable and easier to detect. Moreover, after predicting the building corner vertices, intensive vertex resampling of contours and redundant vertex removal will no longer be required, resulting in the method pipeline being more concise. Meanwhile, the background parts of hollow buildings (as shown in Fig. 14(a)) should be correctly eliminated.

# 5 Conclusion

In this paper, we have proposed BuildMapper, which is an end-to-end, simple, and highly efficient contour-based method for building contour delineation from remote sensing images. Unlike existing methods that rely on extensive post-processing, BuildMapper directly yields vectorized manual-delineation-level building contours via the contour evolution module that simultaneously adjusts the contour vertices and eliminates redundant vertices, achieving the first end-to-end regular vector format building contour extraction. Moreover, the learnable instance-aware contour initialization module is introduced to replace the empirically designed initial contours used in existing contour-based methods, and the proposed dynamic matching loss (DML) solves the optimal pairing problem between the ground-truth vertices and the predicted dynamic contour vertices. The edited large-scale WHU-Mix (vector) dataset will facilitate the advancement of contour-based building extraction methods. The extensive experiments conducted on multiple datasets demonstrated BuildMapper's superiority over the recent methods in the accuracy of the segmentation and boundaries, the proportion of extracted individual building contours that reach the human delineation level, and efficiency. In summary, we believe that this paper not only has methodological innovation, but has also introduced a promising new technique.

## Declaration of Interest

The authors declare that they have no conflicts of interest.

## Acknowledgement

This work was supported by the National Natural Science Foundation of China (grant No. 42171430) and the State Key Program of the National Natural Science Foundation of China (grant No. 42030102).